\documentclass{article} 
\usepackage{gpl_iclr2022_conference,times}

\usepackage{amsmath,amsfonts,bm}

\def\eqref#1{equation~\ref{#1}}

\def\1{\bm{1}}

\def\va{{\bm{a}}}

\def\vs{{\bm{s}}}

\def\vv{{\bm{v}}}

\DeclareMathAlphabet{\mathsfit}{\encodingdefault}{\sfdefault}{m}{sl}
\SetMathAlphabet{\mathsfit}{bold}{\encodingdefault}{\sfdefault}{bx}{n}

\def\sA{{\mathbb{A}}}

\def\sS{{\mathbb{S}}}

\newcommand{\E}{\mathbb{E}}

\newcommand{\R}{\mathbb{R}}

\usepackage{hyperref}
\usepackage{url}
\usepackage{comment}
\usepackage{xcolor}
\usepackage{cancel}
\usepackage{pdfpages}
\usepackage{caption}
\usepackage{subcaption}
\usepackage{svg}
\usepackage{amsmath}
\usepackage{booktabs}
\usepackage{amssymb}
\usepackage{pifont}
\usepackage{tikz}
\usepackage[]{algorithm2e}

\newcommand{\cmark}{\ding{51}}%
\newcommand{\xmark}{\ding{55}}%

\newcommand*{\affmark}[1][*]{\textsuperscript{#1}}

\iclrfinalcopy

\author{
Vlad Sobal\affmark[1], Alfredo Canziani\affmark[2], Nicolas Carion\affmark[2], Kyunghyun Cho\affmark[1, 2, 3, 4], Yann LeCun\affmark[1, 2, 5] \\
\affmark[1] Center for Data Science, New York University \\
\affmark[2] Courant Institute, New York University \\
\affmark[3] Prescient Design, Genentech \\
\affmark[4] CIFAR Fellow \\
\affmark[5] Meta AI Research \\
\texttt{us441@nyu.edu}
}

\newcommand{\senv}{\vs^{\mathrm{env}}}
\newcommand{\sself}{\vs^{\mathrm{self}}}
\newcommand{\aself}{\va^{\mathrm{self}}}

\newcommand{\fenv}{f^{\mathrm{env}}}
\newcommand{\fself}{f^{\mathrm{self}}}

\newcommand{\Senv}{\sS^{\mathrm{env}}}
\newcommand{\Sself}{\sS^{\mathrm{self}}}
\newcommand{\Aself}{\sA^{\mathrm{self}}}

\newcommand{\Mcar}{M^{\mathrm{car}}}
\newcommand{\Mside}{M^{\mathrm{side}}}

\newcommand{\CFM}{CFM}
\newcommand{\CFMKM}{\CFM-KM}
\newcommand{\DFMKM}{DFM-KM}
\newcommand{\PL}{PL}
\newcommand{\MPC}{MPC}
\newcommand{\CFMKMMPC}{\CFMKM{} \MPC}
\newcommand{\CFMP}{\CFM{} \PL}
\newcommand{\CFMKMP}{\CFMKM{} \PL}
\newcommand{\DFMKMMPC}{\DFMKM{} \MPC}

\newcommand{\cproximity}{C_{\mathrm{proximity}}}
\newcommand{\clane}{C_{\mathrm{lane}}}
\newcommand{\coffroad}{C_{\mathrm{offroad}}}

\begin{document}

\title{Separating the World and Ego Models for Self-Driving}

\maketitle

\begin{abstract}
Training self-driving systems to be robust to the long-tail of driving scenarios is a critical problem.
Model-based approaches leverage simulation to emulate a wide range of scenarios without putting users at risk in the real world.
One promising path to faithful simulation is to train a forward model of the world to predict the future states of both the environment and the ego-vehicle given past states and a sequence of actions.
In this paper, we argue that it is beneficial to model the state of the ego-vehicle, which often has simple, predictable and deterministic behavior, separately from the rest of the environment, which is much more complex and highly multimodal.
We propose to model the ego-vehicle using a simple and differentiable kinematic model, while training a stochastic convolutional forward model on raster representations of the state to predict the behavior of the rest of the environment.
We explore several configurations of such decoupled models, and evaluate their performance both with Model Predictive Control (MPC) and direct policy learning.
We test our methods on the task of highway driving and demonstrate lower crash rates and better stability. 
The code is available at
\href{https://github.com/vladisai/pytorch-PPUU/tree/ICLR2022}{https://github.com/vladisai/pytorch-PPUU/tree/ICLR2022}.
\end{abstract}

\section{Introduction}
\label{intro}

Models of the world have proven to be useful for various tasks \citep{Hafner_Lillicrap_Norouzi_Ba_2020, Ebert_Finn_Dasari_Xie_Lee_Levine_2018, Kaiser_Babaeizadeh_Milos_Osinski_Campbell_Czechowski_Erhan_Finn_Kozakowski_Levine_2020}, including self-driving \citep{Henaff_Canziani_LeCun_2019, Ha_Schmidhuber_2018}. 

In their work, \cite{Henaff_Canziani_LeCun_2019} develop a model-based approach to policy learning for highway driving.
The world model in the proposed system is trained to predict
semantic rasterization of the top-down view of a section of a highway around the ego-vehicle, as well as the position and velocity of the ego-vehicle. The policy model interacts with this world model and gets updated by following the
gradient calculated by backpropagation from a handcrafted cost through the world model and into the policy parameters.
This approach allows for training policies without needing additional on-policy data, which is a very important advantage for self-driving.

However, the proposed world model has a few important limitations. The world model needs to predict both the
image of the top-down view and the vehicle state. The two tasks are very different: ego-vehicle state prediction is deterministic and can be computed using a few kinematic equations, while the environment has to be represented by a more complex model capable
of representing multimodal predictions to capture the variety of behaviors of other traffic participants.
This difference suggests the need for two distinct models, not one.
Another limitation of the method proposed by \cite{Henaff_Canziani_LeCun_2019} is the cost function that is non-differentiable with respect to the position of the vehicle.
The gradients flow only through the prediction of the top-down image while omitting important learning signal that can be obtained from the position of the car.

In the present work, we claim that problems that require modeling of agents' motions in a
complex stochastic environment should be addressed with two distinct models: a simple and deterministic
kinematic model that predicts ego-agents' state in the environment, and a complex model of the environment that is
capable of addressing the problem's stochasticity and multimodality. Separating these models provides several advantages.
First, we can leverage the knowledge of kinematics to build an exact ego model.
Second, when the context allows, we can even make the two models completely independent of each other,
making model-predictive control simpler and more efficient.

We build on top of the work of \cite{Henaff_Canziani_LeCun_2019} and make the following contributions:

\begin{itemize}
    \item we introduce a way to split a world model for self-driving into an environment model and an ego model, and compare different ways of integrating them into one system;
    \item we propose a novel approach that makes the two models independent. We demonstrate that such an approach is faster in certain settings while achieving better performance on the task of highway driving;
    \item we design a cost function that is differentiable with respect to the position and velocity of the ego vehicle;
\end{itemize}

\section{Problem description}
\label{problem}

We consider the problem of autonomous highway driving in this paper, although our approach is applicable to controlling any (similar) autonomous system.
We consider three variables at each time step $t$.
They are the self-state $\sself_t \in \Sself$, the self-action $\aself_t \in \Aself$ and the environment state $\senv_t \in \Senv$.
As the names suggest, the first two are the state of and action taken by the autonomous vehicle under control (ego-car), while the environment state includes everything except the ego-car, such as other cars and any other objects and agents.
In our setup $\sself_t \in \R^5, \sself_t = (x_t, y_t, u_t^x, u_t^y, s_t)$, where $(x_t, y_t)$ are coordinates of the center of the rear axle (approximated as the center of the rear end of the car), $(u_t^x, u_t^y)$ is a unit direction vector, and $s_t$ is a scalar denoting the speed. $\aself_t \in \R^2$ is the action taken by the controlled agent at time $t$, with $a_{t, 0}$ and $a_{t, 1}$ denoting the applied acceleration and rotation strength respectively. 
$\senv_t$ is a rasterized mid-level representation of a portion of the highway around the ego-vehicle. An example of such representation is provided in the top row of figure \ref{fig:cost}.
We also assume the availability of a cost function $C : \Senv \times \Sself \times \Aself \rightarrow \R^+$ that, given states and actions at step $t$, calculates the cost. In this work, $C$ has been handcrafted and includes components to account for proximity of other vehicles, driving off the road and closeness to the lane center.
Having all these components, the goal is then to find a policy $\pi$ that minimizes the cumulative cost $J(\pi) = \E_{(\senv_1, \ldots,, \senv_T), (\sself_1, \ldots, \sself_T) \sim \pi}  \left[ \sum_{t=1}^{T} C\left(\senv_t, \sself_t, \pi(\senv_{t-1}, \sself_{t-1})\right)\right]$, where $T$ is episode length.

\paragraph{Dependencies} To create a world model for this problem, we must consider the dependencies among these state and action variables.
We start by exhaustively enumerating these dependencies. State components depend on the state information at the previous time step and the action, while the action depends on the state information:
\begin{itemize}
    \item $\sself_{t+1} \leftarrow \aself_{t}, \sself_{t}, \senv_{t}$
    \item $\senv_{t+1} \leftarrow \aself_{t}, \sself_{t}, \senv_{t}$
    \item $\aself_t \leftarrow \sself_t, \senv_t$
\end{itemize}

The goal of creating a world model then boils down to building a function approximator that predicts $\sself_{t+1}$ and $\senv_{t+1}$ variables given their dependencies.
Once we have the models for the states $(\sself_{t}, \senv_{t})$, we can minimize the cost, or the sum of it over time, w.r.t. the action sequence $(\aself_1, \ldots, \aself_T)$, on which we can train a policy $\pi$ for driving the autonomous vehicle.

\section{World modeling}
\label{world_modeling}

\begin{figure}[t]
    \begin{subfigure}[t]{0.49\linewidth}
         \centering
         \includegraphics[width=\textwidth]{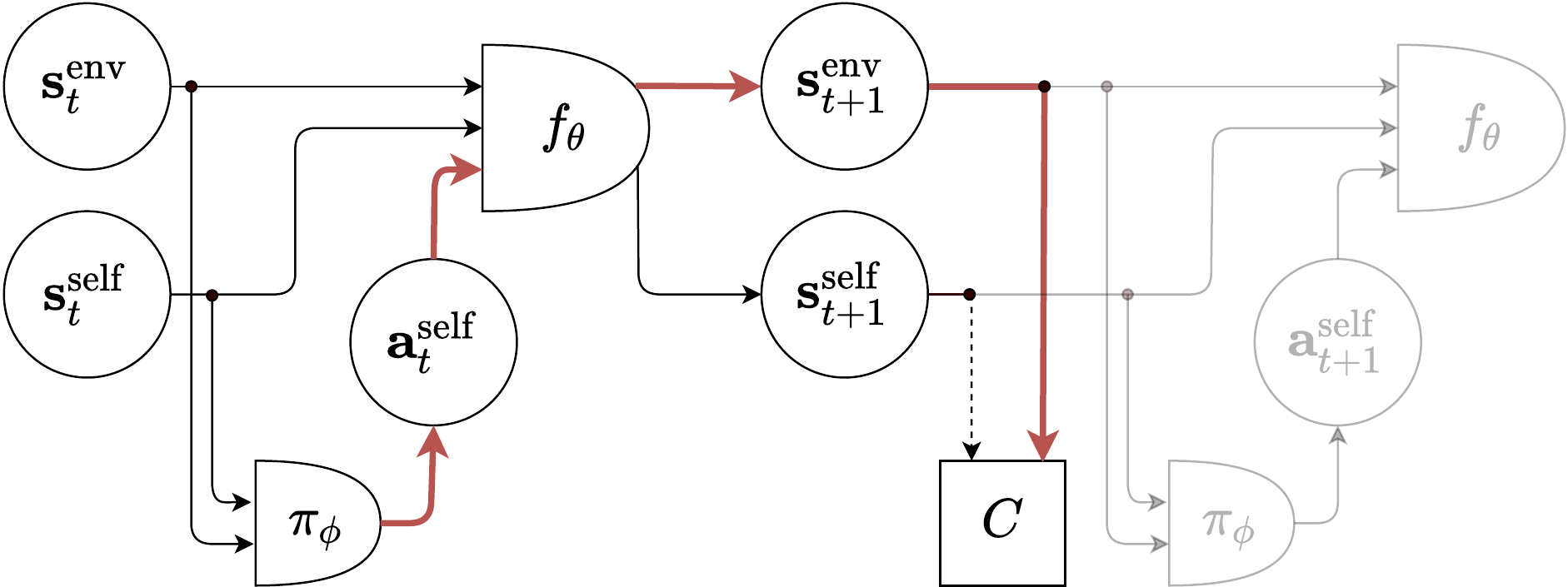}
         \caption{\textbf{CFM Policy} method as proposed by \cite{Henaff_Canziani_LeCun_2019}. There's no
         kinematics model $\fself$. Instead, the forward model $\fenv_\theta$ is tasked with learning
         the kinematics as well as the predicting the trajectories of other agents.}
         \label{fig:CFM_Policy}
     \end{subfigure}
     \hfill
    \begin{subfigure}[t]{0.49\linewidth}
         \centering
         \includegraphics[width=\textwidth]{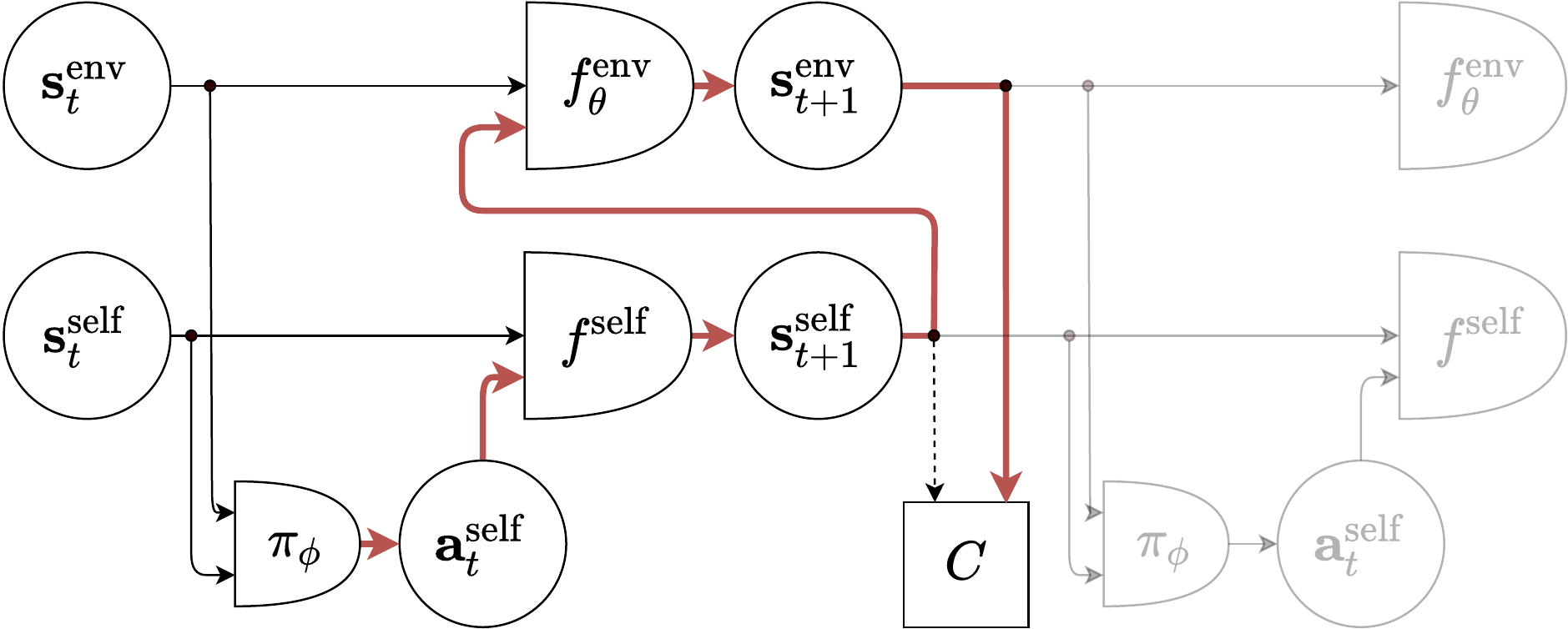}
         \caption{\textbf{CFM-KM Policy.} Here, the forward model does not predict
         $\sself_{t+1}$. The exact kinematics model $\fself$ predicts it instead and passes to $\fenv_\theta$ as
         input.}
         \label{fig:CFM_KM}
     \end{subfigure}
     \par\bigskip 
    \begin{subfigure}[t]{0.49\linewidth}
         \centering
         \includegraphics[width=\textwidth]{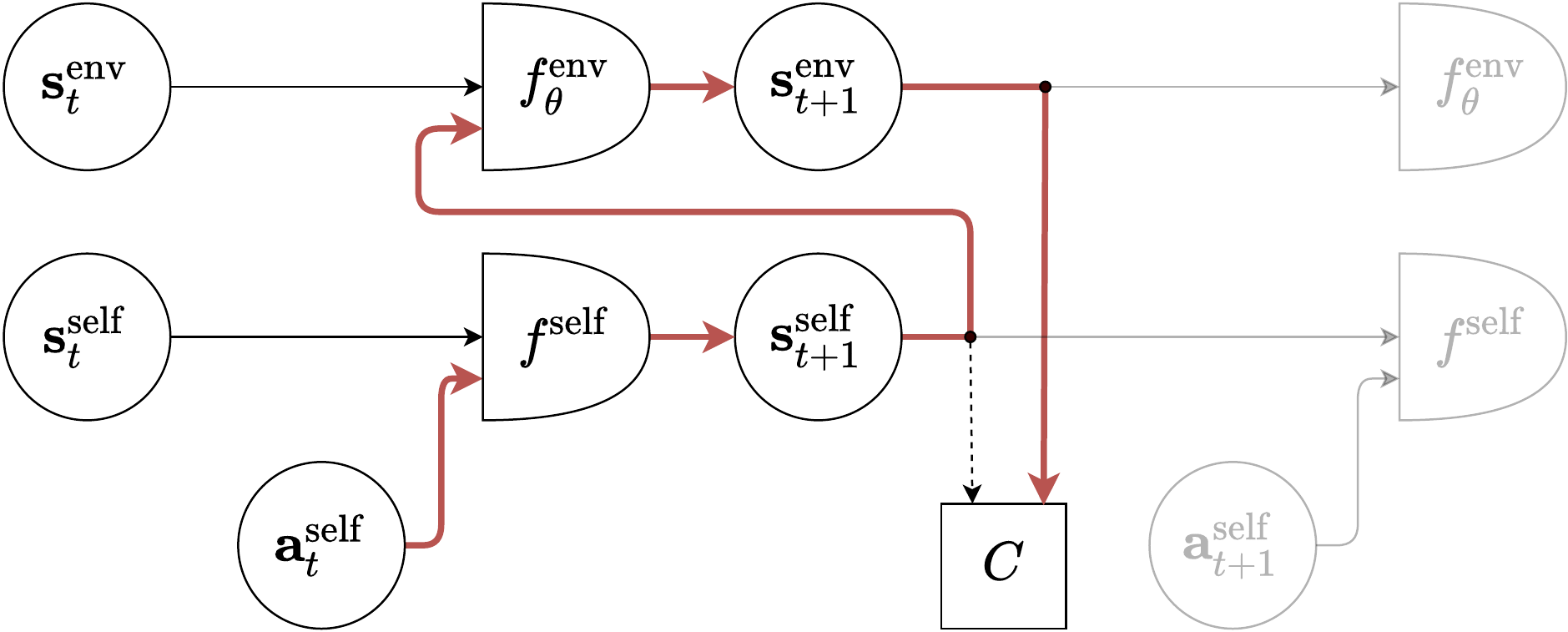}
         \caption{\textbf{CFM-KM MPC.} Like in CFM-KM Policy, we use the kinematics model $\fself$ to help $\fenv$.
         Additionally, instead of learning a policy, we minimize $C$ by directly optimizing the action $\aself_t$ with a few steps of gradient descent.}
         \label{fig:CFM_KM_MPC}
     \end{subfigure}
     \hfill
    \begin{subfigure}[t]{0.49\linewidth}
         \centering
         \includegraphics[width=\textwidth]{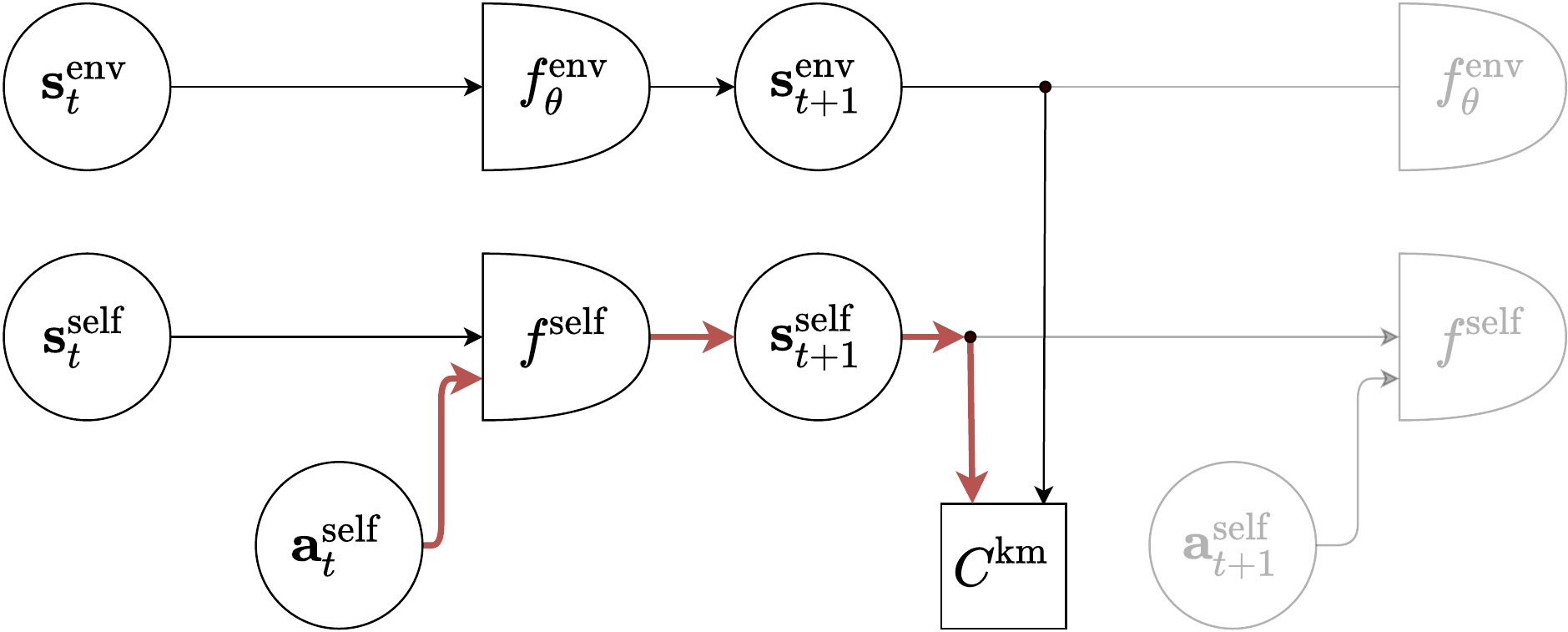}
         \caption{\textbf{DFM-KM MPC.} Here we assume independence of $\fenv_\theta$ from $\sself_t$. The gradients are only propagated through $\fself$. We do not require 
         $\fenv_\theta$ to be differentiable.}
         \label{fig:DFM_KM_MPC}
     \end{subfigure}
    \caption{\textbf{Diagrams of the compared methods \enspace} The circles represent values, while the half ellipses represent functions.
    Arrows represent information flow, and the red color denotes the flow of the gradients.
    Grayed out areas depict the flow into the
    prediction of next time step. 
    Dashed lines depict non-differentiable paths.
    In, CFM-based methods, the cost $C$ is not differentiable w.r.t. $\sself$.} 
    \label{fig:compared_methods}
\end{figure}

In this section, we present the kinematics model and three possible ways of integrating it with the environment model. In section \ref{policy_learning} we present two ways of obtaining a policy using the world model. We then choose 4 combinations of environment model and policy learning set-ups and show experimental results in 
section \ref{experiments}. The 4 combinations we use are shown in figure \ref{fig:compared_methods}. We review related work in section \ref{related_work}, and conclude with section \ref{discussion}.

\paragraph{Kinematics model}
We propose to simplify the dependency pattern of $\sself_t$:
$\sself_{t+1} \leftarrow \aself_{t}, \sself_{t}, {\color{red} \cancel{\senv_{t}}}$. If we 
assume that the state of the environment $\senv_{t}$ does not affect the ego vehicle's state $\sself_{t+1}$, we can resort to a simple bicycle kinematics model for state prediction.
This assumption holds unless there is a collision between the ego car and an object in the environment.

We use the following formulation of the kinematic bicycle model:
\begin{align}
x_{t+1} &= x_{t} + s_t u_t^x \Delta t \\
y_{t+1} &= y_{t} + s_t u_t^y \Delta t \\
s_{t+1} &= s_{t} + a_{t, 0} \Delta t \\
(u_{t+1}^x, u_{t+1}^y) &= \mathrm{unit}[(u_{t}^x, u_{t}^y) + a_{t, 1} \Delta t (u_{t}^y, -u_{t}^x)] \label{new_direction_eq}
\end{align}

Where $\Delta t$ is the time step (we use $\Delta t = 0.1\,\mathrm{s}$), $\mathrm{unit}(\vv) = \frac{\vv}{| \vv |}$. Equation \ref{new_direction_eq} can be intuitively understood as adding to the current 
direction vector an orthogonal unit vector multiplied by the turning command and the time step.

\paragraph{Environment model}

The environment model $\fenv_\theta$ is directly inspired by the work of \cite{Henaff_Canziani_LeCun_2019}.
We use the same architecture in all our experiments with slight changes to the input. 
In all cases, the model $\fenv_\theta$ takes as input $\senv_{t}$ and outputs $\senv_{t+1}$.
Depending on the set-up, the $\fenv_\theta$ may have other inputs and/or outputs.
An example of the prediction of a sequence $\senv_t$ is shown in the top row of figure \ref{fig:cost}. 
To find the best way of integrating the kinematic model into the system, we experiment with three configurations of the environment model:

\emph{1. \enspace Coupled Forward Model (CFM) \quad} is directly taken from \cite{Henaff_Canziani_LeCun_2019}.
In this configuration $\fenv_\theta : \Senv \times \Sself \times \Aself \to \Senv \times \Sself, \;
(\senv_t, \sself_t, \aself_t) \mapsto (\senv_{t+1}, \sself_{t+1})$.
There is no explicit $\fself$ model, instead the world model $\fenv_\theta$ is trained to predict both $\senv_{t+1}$ and $\sself_{t+1}$. The diagram of a set-up using this model is shown in figure \ref{fig:CFM_Policy}. This approach
models all dependencies described in section \ref{problem} with a single model $\fenv_\theta$.

\emph{2. \enspace Coupled Forward Model with Kinematics (CFM-KM) \quad} extends the CFM 
model, but utilizes the proposed kinematic model and the associated independence assumption to better model $\sself_t$. 
We have $\fenv_\theta : \Senv \times \Sself \to \Senv, \;
(\senv_t, \sself_{t+1}) \mapsto \senv_{t+1}$.
The model
of the environment $\fenv_\theta$, instead of taking $\aself_{t}$ as input, takes
the prediction of $\sself_{t+1}$ provided by $\fself(\sself_t, \aself_t)$. Thus, $\fenv_\theta$ does not need to learn the kinematics. Diagrams of two methods using such set-up are shown in figures \ref{fig:CFM_KM} and \ref{fig:CFM_KM_MPC}.

\emph{3. \enspace Decoupled Forward Model with Kinematics (DFM-KM) \quad} In this case, to make the approximation more efficient, we introduce another change to the dependency pattern: $\senv_{t+1} \leftarrow  {\color{red} \cancel{\aself_{t}}, \cancel{\sself_{t}}}, \senv_{t}$. This severs the dependency between the environment and the state of the ego-vehicle. Now, we can run the environment model $\fenv$ separately from $\fself$. 
We then have $\fenv_\theta : \Senv \to \Senv, \;
\senv_t \mapsto \senv_{t+1}$. As before, $\sself_{t+1}$ is predicted using $\fself(\sself_t, \aself_t)$.
A diagram of a set-up using such pattern is shown in figure \ref{fig:DFM_KM_MPC}.

\section{Policy}
\label{policy_learning}

\begin{figure}[h]
     \centering
     \includegraphics[width=0.6\textwidth]{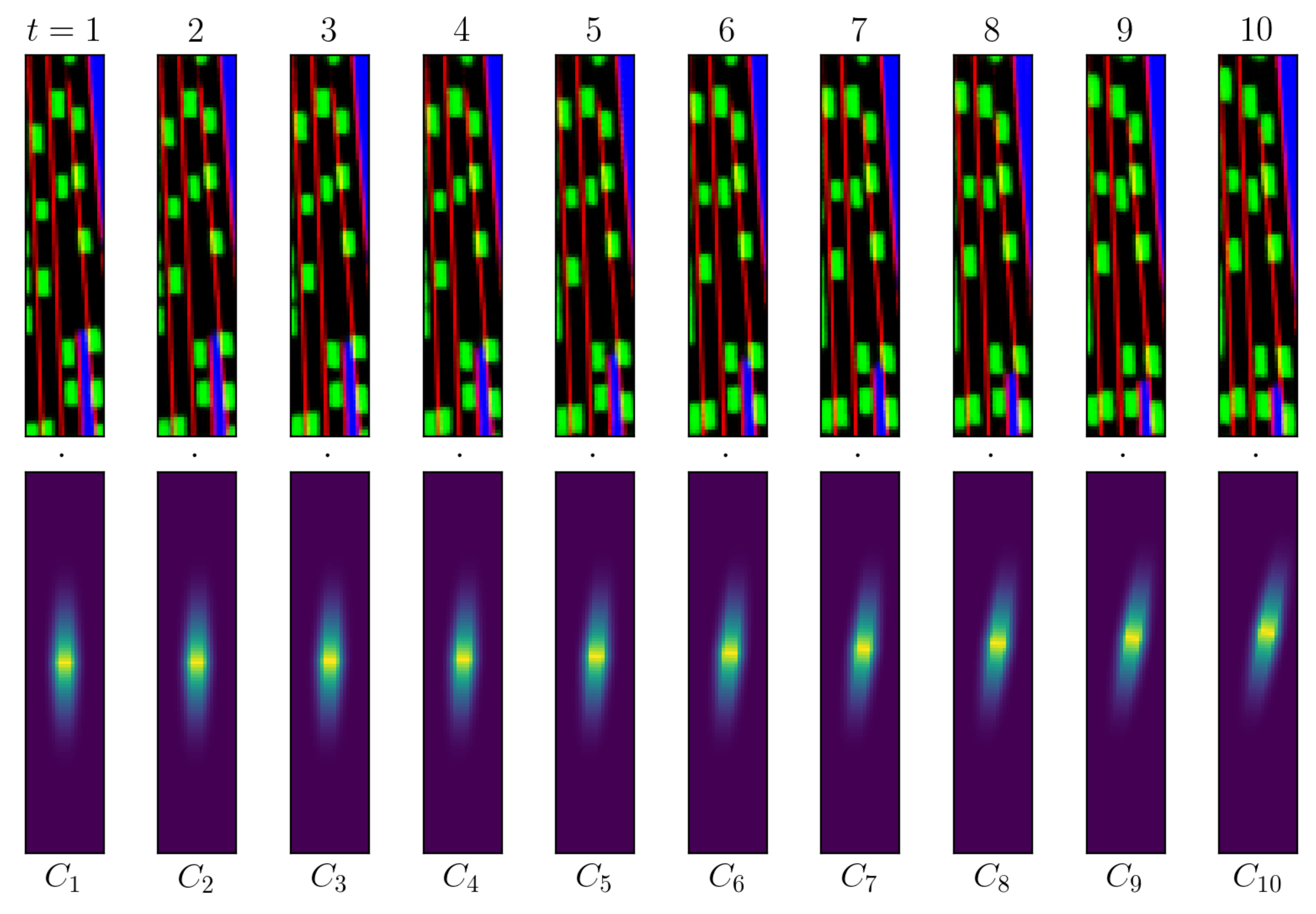}
     \caption{\textbf{Cost calculation.} The top row shows $\fenv_\theta$ predictions for $\senv$ across the time steps. Red, green, and blue denote lane
     demarcations, road users, and off-road regions respectively. The bottom row shows
     masks used for calculating cost components. The masks' location is differentiably adjusted based on $\fself$ prediction of $\sself_{1:10}$. $\cdot$ represents dot product. The bottom row masks are multiplied with individual channels of the corresponding 
     top row images to obtain the values of $C_{1:10}$.
     }
     \label{fig:cost}
\end{figure}

\paragraph{Cost function} CFM PL uses the same cost function as \cite{Henaff_Canziani_LeCun_2019}. \CFMKM{} and \CFMKMMPC{} add one modification: the off-road component. Adding that cost component to CFM PL cost function does not change performance by much, we show results in appendix \ref{sec:offroad_cost}. The resulting cost function $C(\senv_t, \sself_t)$ contains components to account for proximity to other road
users, crossing lane demarcations and driving off the road.
For the \DFMKM{} set-up, we implement $C^\mathrm{km}$, a modification of the original cost function $C$ that 
is differentiable with respect to $\sself$.
In its original version, the cost is used for backpropagation through $\senv_t$ and the forward model $\fenv_\theta$ (see figure \ref{fig:CFM_Policy}).
With the decoupled model, we no longer require the forward
model to be differentiable and instead perform backpropagation through $\sself$. 
To calculate the cost of taking a sequence of actions of length $T$ $(\aself_{1}, \ldots, \aself_{T})$, we first run $\fenv_\theta$ to obtain the predictions of $(\senv_{1}, \ldots, \senv_{T})$ (see the top row of figure \ref{fig:cost}). 
Then, we use $\fself$ to predict $(\sself_{1}, \ldots, \sself_{T})$, which are then used to create masks shifted to locations
that match the sequence of $\sself$ (see the bottom row of figure \ref{fig:cost}). The mask is shifted in a differentiable manner to allow gradient propagation to $\sself$. The masks are then multiplied with individual channels of the predicted $\senv$ 
to obtain different components of the cost, which are then combined into one scalar with corresponding weighting coefficients. For a more in-detail explanation of the cost calculation, see appendix \ref{cost_explanation}. With this method, we can backpropagate into the action sequence $(\aself_{1}, \ldots, \aself_{T})$, 
update it following the negative direction of the gradients, and repeat the whole process until the cost is low enough. 
Note that since we do not backpropagate through $\senv_t$ or $\fenv_\theta$, we do not need to re-run the forward model
at each optimization step.

\paragraph{Policy} 
We utilize two approaches for obtaining the driving policy.

\emph{1. \enspace Model Predictive Control (\MPC) \quad} At step $t=0$, having a sequence of planned actions of length $T$ $(\aself_0, \ldots, \aself_{T-1})$, we want to minimize the cost associated with that plan. We first use forward models $\fenv_\theta$ and $\fself$ to predict $(\senv_1, \ldots, \senv_T)$ and $(\sself_1, \ldots, \sself_T)$, and then use the cost $C$ to obtain the total cost of the predicted trajectory $J = \sum_{t=1}^T C(\senv_t, \sself_t, \aself_{t-1}) \cdot \gamma^t$, where $\gamma$ is the discounting factor that is set to $0.99$.
Assuming that the cost $C$ calculation is differentiable w.r.t. the actions, we can backpropagate the gradients into the sequence of actions (see figure \ref{fig:CFM_KM_MPC} and \ref{fig:DFM_KM_MPC}).
We then do several steps of gradient descent to update the action sequence to minimize the cost.
Having the optimized sequence of actions, we then take the first action $\aself_1$ in the sequence and discard the rest, only to re-plan again at the next time step. For more details, see the pseudocode of DFM-KM MPC and CFM-KM MPC in figure \ref{fig:code} in the appendix.

\emph{2. \enspace Policy Learning (\PL) \quad}
This approach is inspired by the method proposed by \cite{Henaff_Canziani_LeCun_2019}. $\aself_t$ can be modeled using a model of the policy $\pi_\phi(\senv_{t}, \sself_{t})$. We train a policy to minimize the sum of the costs $J = \sum_{t=1}^T C(\senv_{t}, \sself_t, \pi_\phi(\senv_{t-1}, \sself_{t-1})) \cdot \gamma^t$ over a roll-out trajectory of $T$ steps.
This can be done by computing the negative gradient of the sum of the costs w.r.t. the policy's parameters $\phi$ and repeatedly taking the step towards it (see figure \ref{fig:CFM_Policy}). 

In the case of both \MPC{} and \PL, the proposed sparse dependency pattern simplifies backpropagation. Gradients flow only through the self-states $\sself_t$, as we have severed the dependency between the environment state and the self-state. This dramatically lowers the number of interactions involved in the backward pass. This should alleviate the issue of vanishing gradients and improve the quality of the gradients flowing to actions $\aself_t$.

\paragraph{Expected benefits}
We expect the methods proposed in section \ref{world_modeling} to improve the obtained policies in the following ways:
\begin{enumerate}   
    \item We expect the policies to have improved generalization and lower variance of predicted actions when using the kinematics model and/or decoupled forward model. Backpropagating through a simpler and more exact model should enable us to train a more robust policy.
    \item Decoupled forward model simplifies the model that connects the action to 
    the cost. Therefore, we expect each backward pass to take less time, making MPC faster than 
    in the coupled approach.
\end{enumerate}

We test both the existence and degree of these benefits in our experiments.

\paragraph{A potential drawback}

The obvious drawback of the sparsification in the proposed \DFMKM{} is that the environment state and self-state may eventually become incompatible with each other. Because objects in the environment are not aware of the ego-car, some of them may eventually overlap with the ego-car, resulting in an unrealistic situation, such as squeezing in a traffic jam. 
To avoid such unrealistic situations from impacting the policy, we only use a limited roll-out when using the proposed sparse dependency pattern.
We argue that this is fine from two perspectives.
First, even with the conventional dense dependency pattern, learning a policy by backpropagating through a recurrent network, which is how a world model is often implemented, is challenging because of vanishing or exploding gradients.
Therefore, the effective horizon of backpropagation does not decrease much by using a limited roll-out with the sparse dependency pattern.
Second, our task of lane following does not require long-range planning by construction. The policy only needs to repeat short-term goals, \emph{i.e.} to maintain speed and distance from other objects on the road, over and over.

\section{Experiments}
\label{experiments}

\paragraph{Dataset} We test our methods on the task of highway driving on the NGSIM I-80 dataset \citep{ngsim}.
The dataset consists of highway driving scenarios recorded from multiple cameras mounted above a section of Highway I-80 in California.
The recordings take place at different times of day to maximize the diversity of traffic densities.
We follow the pre-processing steps of \cite{Henaff_Canziani_LeCun_2019} and obtain cars' dimensions and trajectories on the highway.
We use the same dataset split of 80\%, 10\%, 10\% for training, validation, and testing respectively.

\paragraph{Crash rates comparison}
We compare the selected combinations of approaches to policy learning and forward modeling proposed in section \ref{world_modeling}. 
The components used by the methods are encoded in the names. For implementation details, refer to appendix \ref{implementation}. The compared methods are:

\emph{(a) \enspace \CFMP \quad} See figure \ref{fig:CFM_Policy}. This is the approach proposed by \cite{Henaff_Canziani_LeCun_2019}.

\emph{(b) \enspace \CFMKMP \quad} See figure \ref{fig:CFM_KM}. This augments \CFMP{} by adding the exact kinematic model following the method described in section \ref{world_modeling}.

\emph{(c) \enspace \CFMKMMPC \quad} See figure \ref{fig:CFM_KM_MPC}. Same as \CFMKMP, but it uses \MPC{} to find the best action.

\emph{(d) \enspace \DFMKMMPC \quad} See figure \ref{fig:DFM_KM_MPC}. As described in section \ref{world_modeling}, this combines \MPC{} with exact kinematic model, decoupled forward model, and the modified cost function.

\begin{table}[t]
\caption{\textbf{Crash rates comparison.} We compare episode failure rates in two simulation setups: replay, where other cars are following the trajectories from the recorded dataset; and interactive simulation, where other cars are controlled either by CFM-KM PL or CFM PL methods. Lower crash rates are better. }
\label{results-perf}
\begin{center}
\begin{tabular}{lrrr}
& &\multicolumn{2}{c}{\bf Interaction Policy}  \vspace{0.1cm} \\
\multicolumn{1}{c}{\bf Method} 
&\multicolumn{1}{c}{\bf Fixed Replay} 
&\multicolumn{1}{c}{\bf CFM-KM PL}
&\multicolumn{1}{c}{\bf CFM PL}
\\ \hline \\
CFM PL  & $25.2 \pm 3.0$          & $9.5 \pm 1.0$          & $5.7 \pm 2.6$ \\
CFM-KM PL  & $15.1 \pm 1.9$          & $\mathbf{1.0 \pm 0.1}$ & $\mathbf{1.1 \pm 0.3}$ \\
CFM-KM MPC     & $25.4 \pm 1.4$          & $4.2 \pm 1.1$          & $3.3 \pm 0.9$ \\
DFM-KM MPC     & $\mathbf{13.2 \pm 1.2}$ & $1.5 \pm 0.5$          & $1.7 \pm 0.6$ \\
\end{tabular}
\end{center}
\end{table}
We test all methods in two settings: replay simulation and interactive simulation.
Replay simulation simply replays the trajectories of all cars except one, which is controlled by the method we are evaluating. 
This is the same evaluation protocol as was used by \cite{Henaff_Canziani_LeCun_2019}.
This evaluation method has a severe limitation: other cars' actions are simply replayed from the 
dataset, and therefore are independent of the ego car's actions.
Some unrealistic situations may happen, for example, the ego-car can be squeezed by other cars in a traffic jam if the ego car picks a different trajectory from the one that the original vehicle followed during data recording.
Since the proposed \DFMKMMPC{} method assumes exactly such independence, replay evaluation results may be biased in 
its favor.
To address this problem, we also show the results of interactive simulation.
Inspired by \citep{Bergamini_Ye_Scheel_Chen_Hu_Del_Pero_Osinski_Grimmett_Ondruska_2021}, we implement interactive simulation by controlling the ego-car with the selected method and controlling all other cars with either the \CFMKMP{} or the \CFMP.
We do not experiment with controlling other cars with \MPC{} because it is orders of magnitude slower than doing one forward pass with a policy model (see table \ref{results-time}), rendering such evaluation setup too slow to be practical.
The results are shown in table \ref{results-perf}. DFM-KM MPC that uses the decoupled forward model with kinematics achieves the best performance in replay setting, closely trailed by the CFM-KM PL policy learned with the enhanced coupled forward model. 
This suggests that augmenting the forward model with exact kinematics equations gives a great boost in performance compared to the model that has to learn the kinematics from data. In the interactive setting, the
CFM-KM PL performs slightly better, meaning that the independence assumption indeed biases
the results somewhat in the replay evaluation in favor of DFM-KM MPC. However, DFM-KM MPC outperforms CFM-KM MPC by a big margin in all settings, showing
that propagating the gradients through $\sself$ and decoupling the forward model indeed improves the gradients'
quality, allowing MPC to efficiently find better actions.

\paragraph{Time performance} 
Another benefit of the decoupled approach is the improved speed performance of \MPC.
To demonstrate that, we measure the average time needed to evaluate one time step, and show the results in table \ref{results-time}.
DFM-KM MPC needs about half the amount of time needed for the CFM-KM MPC. However,
it is still orders of magnitude slower than running a trained policy. 
\paragraph{Variance of actions}
We also test if the sparse dependency pattern facilitates more robust predictions by the policies. We show results in table \ref{results-time}. We observe that introducing the kinematic model helps to
make the behavior more robust with respect to random initialization, with additional improvement gained from
applying the proposed \DFMKM{} method. For an in-detail explanation of how these variances were computed, see appendix \ref{variance_explanation}.

\begin{figure}[t]
    \begin{subfigure}[t]{0.24\linewidth}
         \centering
         \includegraphics[width=0.6\textwidth]{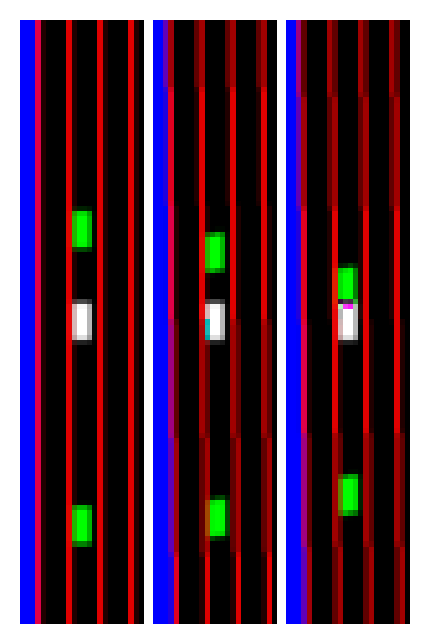}
         \caption{\textbf{\CFMP}}
         \label{fig:CFM_Policy_stress}
     \end{subfigure}
     \hfill
    \begin{subfigure}[t]{0.24\linewidth}
         \centering
         \includegraphics[width=0.6\textwidth]{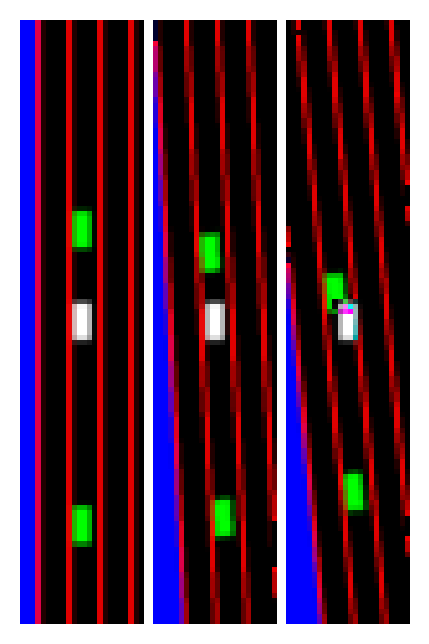}
         \caption{\textbf{\CFMKMP}}
         \label{fig:CFM_KM_stress}
     \end{subfigure}
     \hfill
    \begin{subfigure}[t]{0.24\linewidth}
         \centering
         \includegraphics[width=0.6\textwidth]{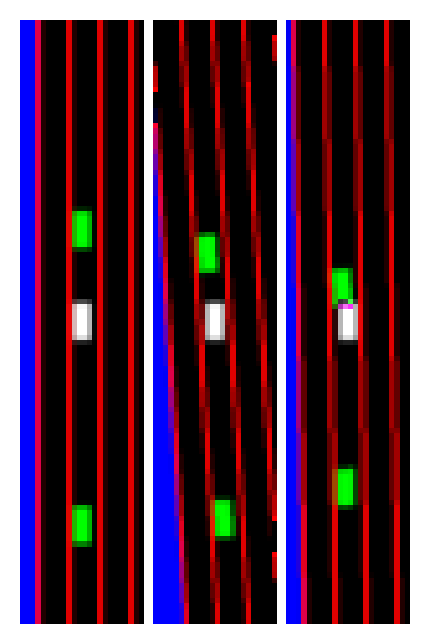}
         \caption{\textbf{\CFMKMMPC}}
         \label{fig:CFM_KM_MPC_stress}
     \end{subfigure}
     \hfill
    \begin{subfigure}[t]{0.24\linewidth}
         \centering
         \includegraphics[width=0.6\textwidth]{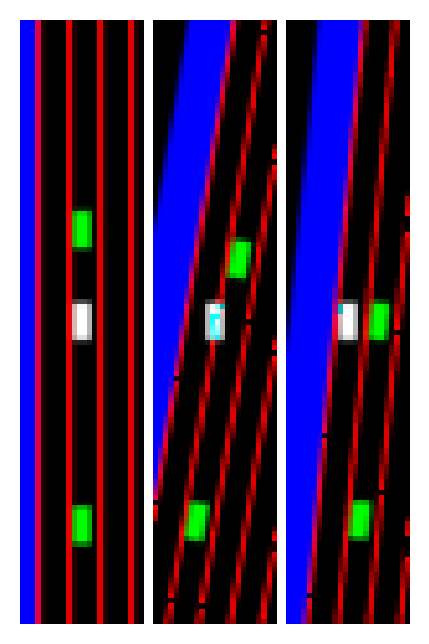}
         \caption{\textbf{\DFMKMMPC}}
         \label{fig:DFM_KM_MPC_stress}
     \end{subfigure}
    \caption{\textbf{Stress-testing the proposed methods.} The controlled car (white, in the center)
    is cruising between two other cars (green) when
    the car directly in front brakes suddenly.
    DFM-KM MPC manages to react in time, while other methods fail.
    \label{fig:stress}
    }
    
\end{figure}

\begin{table}[t]
\caption{\textbf{Time performance and output variance}. To measure agreement among policies using the same method but different seeds, we run the policies on the same input and calculate standard deviation of the produced actions.}
\label{results-time}
\begin{center}
\begin{tabular}{lrrrr}
& & \multicolumn{3}{c}{\bf Standard deviation across seeds}  \vspace{0.2cm} \\
\multicolumn{1}{c}{\bf Method}  
&\multicolumn{1}{p{3cm}}{\centering \bf Milliseconds per \\ simulation step} 
&\multicolumn{1}{c}{\bf Acceleration} 
&\multicolumn{1}{c}{\bf Turning}
&\multicolumn{1}{c}{\bf Average}
\\ \hline \\
CFM PL              & $\bm{1.2 \pm \textcolor{white}{0}0.0}$ & $1.08$ & $1.00$ & $1.04$ \\
CFM-KM PL           & $\bm{1.2 \pm \textcolor{white}{0}0.0} $ & $1.07$ & $\bm{0.70}$ & $0.88$\\
CFM-KM MPC          & $1162.6 \pm 34.9 $  & $1.05$ & $0.77$ & $0.91$ \\
DFM-KM MPC          & $509.5 \pm 18.5 $  & $\bm{0.78}$ & $0.83$ & $\bm{0.80}$ \\
\end{tabular}
\end{center}
\end{table}
\paragraph{Stress-testing} We test the proposed methods on a
hand-designed scenario --- controlling an agent on a highway while cruising between two cars, when the car directly in the front 
brakes suddenly. We show the results in figure \ref{fig:stress}. We observe that only \DFMKMMPC{} method manages to successfully complete the scenario. CFM-KM MPC fails, showing that backpropagating through $\fenv_\theta$ is 
not as efficient as backpropagating through $\fself$ in DFM-KM MPC approach. We hypothesize that two factors 
help \DFMKMMPC{} here. First, in such extreme scenarios the decoupled model has an advantage since it is unreasonable to expect that the car
in the front will react to the ego-agent. The independence assumption is justified, and helps the optimization process to find the best
action. 
Second, a learned policy, trained to solve multiple different scenarios, likely becomes a smooth function that cannot take
extreme values, while \MPC{} is not restrained by model capacity and can find the minimum of the cost function $C$ better in such extreme cases. 

\section{Related Work}

\label{related_work}
Our work is related to the research on model-predictive control and motion planning methods. These are areas with
decades of research; for comprehensive overviews of these topics we refer the reader to the books on optimal control \citep{Bryson_Ho_AOC, Bert05}, and motion planning \citep{LaValle_2006}. In this section, we mainly focus on the recent
methods that combine the existing approaches with deep neural networks.

\paragraph{Motion planning and Trajectory following}  \cite{Paden_Cap_Yong_Yershov_Frazzoli_2016} provide a survey of the existing methods for self-driving.
Approaches range from graph-search methods, such as A* \citep{Ziegler2008NavigatingCR}, to dynamic programming \citep{Montemerlo_Becker_Bhat_Dahlkamp_Dolgov_Ettinger_Haehnel_Hilden_Hoffmann_Huhnke_etal_2008}. In our work, trajectory following is simplified as we are acting in a simulator with a perfectly controllable car, and 
we focus on motion planning instead.

\paragraph{World modeling} has proved to be a great approach to many tasks due to superior sample complexity in policy learning \citep{Nagabandi_Kahn_Fearing_Levine_2017}, although it comes with some caveats, such as the danger of compounding errors \citep{Asadi_Misra_Kim_Littman_2019}, and stochasticity \citep{Denton_Fergus_2018}.
World model used in our work combines ideas of action-conditioned model \citep{Oh_Guo_Lee_Lewis_Singh_2015}, and stochastic video prediction \citep{Babaeizadeh_Finn_Erhan_Campbell_Levine_2018}. Such models have also been used in \citep{Hafner_Lillicrap_Norouzi_Ba_2020} and \citep{Henaff_Zhao_LeCun_2017}.

\paragraph{Model Predictive Control} has been used widely for self-driving. \cite{Zhang_Liniger_Borrelli_2018} use \MPC{} for planning and collision avoidance, \cite{Drews_Williams_Goldfain_Theodorou_Rehg_2018} use \MPC{} to 
build an impressive system that drives a scaled-down vehicle at high speeds around a track.
Our approach can be viewed as a type of Stochastic MPC \citep{Heirung_Paulson_OLeary_Mesbah_2018}, or Scenario-based MPC \citep{Schildbach_Fagiano_Frei_Morari_2014, Cesari_Schildbach_Carvalho_Borrelli_2017}, where the trajectory is optimized for a limited number of future scenarios (in our case this number is 1, but it can easily be increased).

\paragraph{Kinematic models} have been used extensively in self-driving applications.  Often, cars are
approximated by simplified models, such as unicycle \citep{predictionnet} or bicycle models \citep{Cesari_Schildbach_Carvalho_Borrelli_2017}. These models
are particularly useful when adding inductive bias to models to produce realistic trajectories in
path planning or behavior prediction \citep{trajectronpp}. \cite{mpckmtraj}  provide an overview of kinematics and dynamics models used for self-driving and apply them to path-following. The work of \cite{urban_driver} is particularly close to
ours as they also propose a differentiable kinematic model for training a self-driving policy. However, there is no trained environment model, and the predictions of the other road users are replaced with log replay.

\paragraph{Interactive simulation} is a long-standing problem in self-driving cars development. \cite{Bergamini_Ye_Scheel_Chen_Hu_Del_Pero_Osinski_Grimmett_Ondruska_2021} proposed
a method that uses GANs \citep{Goodfellow_Pouget-Abadie_Mirza_Xu_Warde-Farley_Ozair_Courville_Bengio_2014} to generate the initial state, and then sequentially apply a learned policy to each of the generated agents.
\cite{Suo_Regalado_Casas_Urtasun_2021} propose a system that models the agents' behavior jointly, making more consistent predictions.

\section{Discussion \& Conclusion}
\label{discussion}

We presented a novel design of a world model for an agent with known kinematics in a complex stochastic environment.
The conducted experiments show that the separation of the world model into an ego model and a model of the environment helps
obtain policies that reach better performance in our experiments with highway driving. Decoupling these two models
completely makes MPC perform faster and better, and helps to solve a stress-test scenario that requires quick reaction from the policy. We believe that our proposed approach can be applied to any problem that involves an agent with 
easily predictable kinematics acting in a complex stochastic environment, e.g. controlling robots in the real world, such as delivery carts or drones; or some Atari games, such as Space Invaders, or Freeway.
We also believe that for the suggested separation to work, it is not strictly necessary to have the exact kinematic equations, the ego model can also be learned.

There is still more to explore about the proposed approaches. First, although DFM-KM MPC performs better than CFM-KM MPC, it is yet unclear if that is because of the modified cost function $C^\mathrm{km}$, or because of the decoupled forward model. Experiments with a method that integrates $C^{km}$ with CFM-KM PL would resolve this ambiguity. Second, comparing the results of CFM-KM PL and CFM-KM MPC, we see that policy learning
performs much better, suggesting that DFM-KM PL is an approach worth investigating. 
Third, hand-designing the cost function is only possible for simple contexts, such as highway driving.
For applications to more complex scenarios like urban driving, we would need to learn the cost function, which is highly non-trivial \citep{Ng00algorithmsfor}.

\section{Acknowledgements}
This material is based upon work supported by the National Science Foundation under NSF Award 1922658.

\bibliography{main}
\bibliographystyle{gpl_iclr2022_conference}

\appendix
\section{Calculating the policy output agreement across seeds}
\label{variance_explanation}
To measure the agreement across seeds in Table \ref{results-time}, we first take three policies that use the
same method but different seeds and run them on 1000 examples from the dataset. We follow the procedure of \cite{Henaff_Canziani_LeCun_2019} and, before unnormalizing models' output, clamp the values to the range $[-3, 3]$.
For each method separately, we then calculate the mean and variance for the actions taken across seeds and data examples to obtain $\mu$ and $\sigma^2$.
We then normalize the actions using these values.
Now, for the normalized values, we calculate the standard deviation across outputs for different seeds for each dataset example separately.
The values are then averaged across the entire 1000 examples to obtain the values reported in Table \ref{results-time}. 
Such a procedure accounts for the fact different methods output values of different magnitudes and avoids skewing the standard deviation comparison.

\section{Implementation details}
\label{implementation}
\paragraph{Forward models}
To train updated forward models, we follow the procedure proposed by \cite{Henaff_Canziani_LeCun_2019}. The only
change to the forward models is the change to the input dimension. The training method was unchanged.

\paragraph{CFM-KM Policy Training}
We use the same model as was proposed by \cite{Henaff_Canziani_LeCun_2019}, and train for 70\,k steps, with batch size 10, and learning rate of 0.0001. We decrease the learning rate by a factor of 10 after 70\% of training.

\paragraph{CFM-KM MPC}
To find the optimal action, we perform gradient descent for 11 iterations, with learning rate of 0.31. 
The cost function is calculated as: $C = C_{\mathrm{proximity}} + 0.32 \cdot C_{\mathrm{lane}} + 0.32 \cdot C_{\mathrm{offroad}}$. The uncertainty cost proposed in \cite{Henaff_Canziani_LeCun_2019} was not used for CFM-KM MPC 
as it made each iteration impractically slow. We used the plan length of 20 frames, which corresponds to 2 seconds.
The hyperparameters were found with random search. We provide pseudo-code in figure \ref{fig:code_CFM_KM_MPC}.

\paragraph{DFM-KM MPC}
To find the optimal action, we perform gradient descent for 27 iterations, with learning rate of 0.48. 
The cost function is calculated as: $C = 91.2 \cdot  C_{\mathrm{proximity}} + 2.88 \cdot C_{\mathrm{offroad}} + 3.06 \cdot C_{\mathrm{lane}} + 0.1 \cdot C_\mathrm{jerk} +0.001 \cdot C_\mathrm{destination}$. We use the plan size of 30.
The hyperparameters were found with random search. We provide pseudo-code in figure \ref{fig:code_DFM_KM_MPC}.

We do not re-train the forward model for this setup, we simply use the CFM-KM, but we always run it with 0-actions --- we get predictions $\senv_{t+1}$ that
correspond to what the world would have looked like if the ego-vehicle kept going at current speed and did not turn. We then assume that changing the actions would not have changed $\senv_{t+1}$.

\begin{figure}
\begin{subfigure}[t]{0.51\linewidth}
\begin{algorithm}[H]
 \KwIn{Models $\fself$ and $\fenv_\theta$ \newline 
      cost function $C^{km}$ \newline
      states $\sself_t$ and $\senv_t$ \newline 
      planning horizon $T$ \newline 
      learning rate $\alpha$ \newline
     number of iterations $N$ }
 \KwOut{Action to be taken at time $t$}
 $\aself_{t:t+T-1} \leftarrow 0$\;
 \For{$k \leftarrow 1$ \KwTo $T$}{
     $\senv_{t+k} \leftarrow \fenv_\theta(\senv_{t+k-1})$\;
 }
 \For{$i\leftarrow 1$ \KwTo $N$}{
    \For{$k \leftarrow 1$ \KwTo $T$}{
        $\sself_{t+k} \leftarrow \fself(\sself_{t+k-1}, \aself_{t+k-1})$\;
    }
    $J \leftarrow \sum_{k=1}^{T} \gamma^k C^\mathrm{km}(\senv_{t+k}, \sself_{t+k}, \aself_{t+k-1})$ \;
    $\aself_{t:t+T-1} \leftarrow \aself - \alpha \frac{\partial J}{\partial \aself_{t:t+T-1}}$ \;
 }
 \KwRet{$\aself_t$}\;
\end{algorithm}
\caption{DFM-KM MPC}
\label{fig:code_DFM_KM_MPC}
\end{subfigure}
\begin{subfigure}[t]{0.51\linewidth}
\begin{algorithm}[H]
 \KwIn{Models $\fself$ and $\fenv_\theta$ \newline 
      cost function $C$ \newline
      states $\sself_t$ and $\senv_t$ \newline 
      planning horizon $T$ \newline 
      learning rate $\alpha$ \newline
     number of iterations $N$ }
 \KwOut{Action to be taken at time $t$}
 $\aself_{t:t+T-1} \leftarrow 0$\;
 \For{$i\leftarrow 1$ \KwTo $N$}{
    \For{$k \leftarrow 1$ \KwTo $T$}{
        $\sself_{t+k} \leftarrow \fself(\sself_{t+k-1}, \aself_{t+k-1})$\;
        $\senv_{t+k} \leftarrow \fenv_\theta(\senv_{t+k-1}, \sself_{t+k-1})$\;
    }
    $J \leftarrow \sum_{k=1}^{T} \gamma^k C^\mathrm{km}(\senv_{t+k}, \sself_{t+k}, \aself_{t+k-1})$ \;
    $\aself_{t:t+T-1} \leftarrow \aself - \alpha \frac{\partial J}{\partial \aself_{t:t+T-1}}$ \;
 }
 \KwRet{$\aself_t$}\;
\end{algorithm}
\caption{CFM-KM MPC}
\label{fig:code_CFM_KM_MPC}
\end{subfigure}
\caption{Algorithms of the proposed MPC methods. Note that DFM-KM-MPC runs $\fenv_\theta$ outside the main
optimization loop, while CFM-KM-MPC runs it inside, causing it to take more time per iteration.}
\label{fig:code}

\end{figure}
\section{Offroad cost}
\label{sec:offroad_cost}
Another important difference between CFM PL and the other methods is that it does not use offroad cost $\coffroad$. The offroad
cost component was introduced to prevent the ego-car from driving off the road. Without it, the
cost of driving off the highway is the same as the cost of simply crossing a lane marker. To understand how much
the offroad cost contributed to the improvement in performance, we test the CFM PL method with offroad cost. The results are
presented in table \ref{offroad_results}. We see some improvement with adding offroad cost, particularly in interactive evaluation, but the performance does not reach the results of CFM-KM PL and DFM-KM MPC.

\begin{table}[h]
\caption{Comparison of CFM PL with and without offroad cost component.}
\label{offroad_results}
\begin{center}
\begin{tabular}{lrrr}
& &\multicolumn{2}{c}{\bf Interaction Policy}  \vspace{0.1cm} \\
\multicolumn{1}{c}{\bf Method} 
&\multicolumn{1}{c}{\bf Fixed Replay} 
&\multicolumn{1}{c}{\bf CFM-KM Policy}
&\multicolumn{1}{c}{\bf CFM Policy}
\\ \hline \\
CFM Policy                  & $25.2 \pm 3.0$ &   $7.3 \pm 2.1$ & $4.8 \pm 2.3$ \\
CFM Policy with $\coffroad$ & $25.4 \pm 1.6$ &  $2.1 \pm 0.8$ & $3.3 \pm 0.2$ \\
\end{tabular}
\end{center}
\end{table}

\section{DFM-KM MPC cost}
\label{cost_explanation}
The cost calculation consists of two stages, as described in Section \ref{policy_learning}: mask creation and cost calculation. 
\paragraph{Mask creation}
We create two kinds of masks: one for proximity cost, and the other for offroad and lane costs. The masks are 
created in a way to align with the predicted center of the car and face the direction of the car's heading. To create masks given
the position relative to the center of the image, direction, speed, width, and length of the agent $(x, y, u_x, u_y, s, w, l)$ we follow the steps below:
\begin{enumerate}
    \item Create a mesh grid of coordinates. The rasterized image resolution we use is 117 by 24, and the size of the corresponding area is 72.2 by 14.8 meters. We first create a matrix of coordinates of each of the cells in the image with respect to the center of the image. $A : \R^{117 \times 24 \times 2}$, $A_{i,j} = [(72.2 / 117 \cdot i - 36.1)), (14.8 / 24 \cdot j - 12.4)]$. 
    \item In order to align the coordinates with the ego-car, we shift and rotate them: $B_{i, j} = R^{u_x, u_y}[A_{i, j, 1} - x, A_{i, j, 2} - y]$, where $R^{u_x, u_y}$ is the rotation matrix for the angle specified by $(u_x, u_y)$.
    \item Construct the masks. First, we define
    the safety distance along the direction of movement: $d_x = 1.5 \cdot (\max(10, s) + l) + 1$,
    and in the orthogonal direction: $d_y = w / 2 + 3.7$. $3.7$ here is the lane width.
    These are the distances beyond which the objects are not taken into account in the cost. Then, we want
    to build a mask that reaches 0 at $d_x$ in the front and behind, at $d_y$ at the sides, and reaches 1 at the car edges.
    The masks are:
    \begin{align}
    M^\mathrm{car}_{i, j} &= \left[ \left( \frac{d_x - |B_{i, j, 1}|}{d_x - l/2} \right)^+\cdot \min \left( \left( \frac{d_y - |B_{i, j, 2}|}{d_y - w/2} \right)^+, 1\right) \right] ^ \alpha \\ 
    M^\mathrm{side}_{i, j} &= \left[ \left( \frac{d_x - |B_{i, j, 1}|}{d_x - l/2} \right)^+ \cdot \left( \frac{d_y - |B_{i, j, 2}|}{d_y - w/2} \right)^+ \right] ^ \alpha
    \end{align}
    The difference between $\Mside$ and $\Mcar$ are in the clamping values above 1: when calculating the cost component accounting for proximity to other vehicles, we would like the mask profile to have a ``flat nose''.
    $\alpha$ is a hyperparameter used to make the mask non-linear. Higher values make the cost grow more rapidly as objects come closer to the ego-vehicle.
\end{enumerate}

Note that all operations are differentiable with respect to $(x, y, u_x, u_y, s)$, allowing us to backpropagate through
$\sself$. 

\begin{figure}[h]
    \begin{subfigure}[t]{0.15\linewidth}
         \centering
         \includegraphics[width=0.7\textwidth]{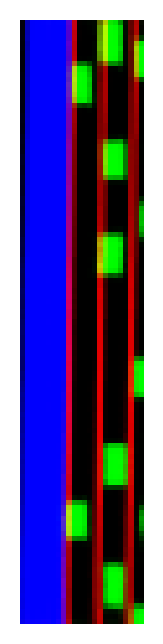}
         \caption{\textbf{$\senv$}}
         \label{fig:together}
     \end{subfigure}
     \hfill
    \begin{subfigure}[t]{0.15\linewidth}
         \centering
         \includegraphics[width=0.7\textwidth]{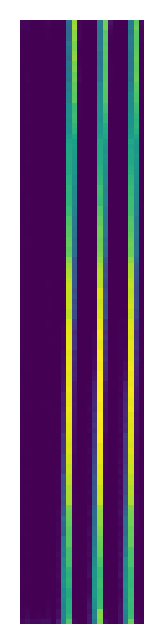}
         \caption{\textbf{$\senv_\mathrm{lanes}$}}
         \label{fig:lanes}
     \end{subfigure}
     \hfill
    \begin{subfigure}[t]{0.15\linewidth}
         \centering
         \includegraphics[width=0.7\textwidth]{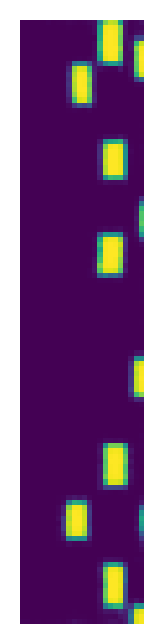}
         \caption{\textbf{$\senv_\mathrm{cars}$}}
         \label{fig:cars}
     \end{subfigure}
     \hfill
    \begin{subfigure}[t]{0.15\linewidth}
         \centering
         \includegraphics[width=0.7\textwidth]{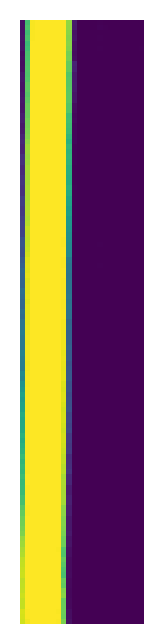}
         \caption{\textbf{$\senv_\mathrm{offroad}$}}
         \label{fig:offroad}
     \end{subfigure}
     \hfill
    \begin{subfigure}[t]{0.15\linewidth}
         \centering
         \includegraphics[width=0.7\textwidth]{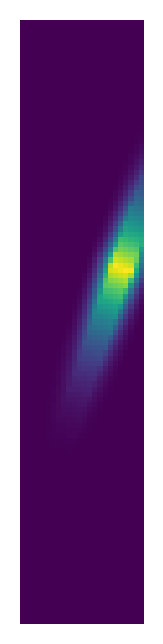}
         \caption{\textbf{$\Mcar$}}
         \label{fig:mcar}
     \end{subfigure}
     \hfill
    \begin{subfigure}[t]{0.15\linewidth}
         \centering
         \includegraphics[width=0.7\textwidth]{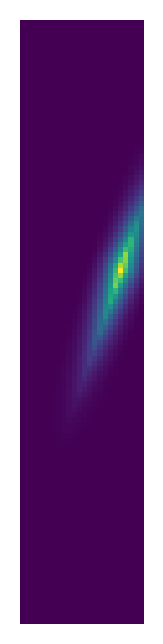}
         \caption{\textbf{$\Mside$}}
         \label{fig:mside}
     \end{subfigure}
    \caption{\textbf{Components used for calulating costs.} Figure \ref{fig:together} depicts the original image, 
    and \ref{fig:lanes}, \ref{fig:cars}, \ref{fig:offroad} depict channels used for cost calculation. \ref{fig:mcar} shows the mask used for car proximity cost, while \ref{fig:mside} is used for offroad and lane costs respectively.
    }
    \label{fig:cost_explanation}
\end{figure}

\paragraph{Calculating cost components} Having the masks, we simply perform element-wise multiplication with the 
corresponding channels of the $\senv$, see figure \ref{fig:cost_explanation}.
\begin{align}
    \clane &= \langle \senv_\mathrm{lanes}, \Mside \rangle \\
    \coffroad &= \langle \senv_\mathrm{offroad}, \Mside \rangle \\
    \cproximity &= \langle \senv_\mathrm{car}, \Mcar \rangle \\
\end{align}

The new cost also introduces two new cost components: destination cost, and jerk cost. 
Destination cost is simply a term that pushes the car to go forward. This prevents cases
where MPC fails to drive forward because there are no cars behind. The cost is calculated as $C_\mathrm{destination} = -x$.
Jerk cost is responsible for making the actions more smooth. Intuitively, this is a cost that penalizes
the derivative of the actions w.r.t. time. $C_\mathrm{jerk} = \frac{1}{T} \sum_{t=2}^T(\aself_t - \aself_{t-1})^\top (\aself_t - \aself_{t-1})$.

The components are then combined into a single scalar with the corresponding weights $\alpha$: 
\begin{align}
\begin{split}
C = & \alpha_\mathrm{lane}\clane +  \\
& \alpha_\mathrm{offroad}\coffroad + \\
& \alpha_\mathrm{proximity}\cproximity + \\
& \alpha_\mathrm{destination}C_\mathrm{destination} + \\
& \alpha_\mathrm{jerk}C_\mathrm{jerk} 
\end{split}
\end{align}

\section{Comparison of methods}

In table \ref{compared_methods} we show the proposed methods and the comparison of the used components.

\begin{table}
\caption{Comparison of characteristics of the tested methods.}
\label{compared_methods}
\begin{center}
\begin{tabular}{lcccc}
\multicolumn{1}{c}{\bf Method} 
&\multicolumn{1}{p{2cm}}{\centering \bf Decoupled Forward Model} 
&\multicolumn{1}{p{2cm}}{\centering \bf Kinematics Model} 
&\multicolumn{1}{p{2cm}}{\centering \bf Learned Policy} 
&\multicolumn{1}{p{2cm}}{\centering \bf Modified cost} 
\\ \hline \\
CFM Policy    & \xmark & \xmark & \cmark & \xmark \\
CFM-KM Policy & \xmark & \cmark & \cmark & \xmark \\
CFM-KM MPC    & \xmark & \cmark & \xmark & \xmark \\
DFM-KM MPC    & \cmark & \cmark & \xmark & \cmark  \\
\end{tabular}
\end{center}
\end{table}


\end{document}